\documentclass[preprints,article,accept,moreauthors,pdftex]{Definitions/mdpi}

\firstpage{1}
\pubvolume{13}
\issuenum{5}
\articlenumber{118}
\pubyear{2020}
\copyrightyear{2020}
\history{Received: 8 April 2020; Accepted: 30 April 2020; Published: 7 May 2020}
\updates{1994} 

\pdfoutput=1
\usepackage{algorithmic}
\usepackage[lined,boxed,ruled,commentsnumbered]{algorithm2e}
\usepackage[utf8]{inputenc}
\usepackage[english]{babel}
\usepackage{amssymb, amsfonts}
\usepackage{braket}
\usepackage{scalerel}
\usepackage{mathtools}
\usepackage{empheq}
\usepackage{mathrsfs}
\usepackage{stmaryrd}
\usepackage{float}
\usepackage{color}
\usepackage{bm}
\usepackage[autolanguage]{numprint}
\usepackage{hyperref}

\renewcommand{\P}{\mathbb{P}}

\newcommand{\R}{\mathbb{R}}

\newcommand{\E}{\mathbb{E}}

\newcommand{\cA}{\mathcal{A}}
\newcommand{\cB}{\mathcal{B}}

\newcommand{\cD}{\mathcal{D}}

\newcommand{\cP}{\mathcal{P}}

\newcommand{\cT}{\mathcal{T}}

\newcommand{\cX}{\mathcal{X}}

\newcommand{\sP}{\mathscr{P}}

\newcommand{\teta}{\widetilde{\teta}}

\newcommand{\e}{\varepsilon}

\renewcommand{\=}{ \coloneqq }

\newcommand{\whQ}{\widehat{Q}}
\newcommand{\wheta}{\widehat{\eta}}
\newcommand{\whcD}{\widehat{\cD}}

\DeclareMathOperator*{\argmax}{arg\,max}

\DeclareMathOperator{\KL}{KL}
\newcommand{\btheta}{\bm{\theta}}
\newcommand{\algnamn}{ECC}
\newcommand{\npr}[2]{\numprint{#1} $\pm$ \numprint{#2}}

\theoremstyle{definition}

\Title{\textls[-20]{Distributional Reinforcement Learning with~Ensembles}}

\Author{Björn Lindenberg * \href{https://orcid.org/0000-0003-2756-5186}{\orcidicon}, Jonas Nordqvist \href{https://orcid.org/0000-0002-0510-6782}{\orcidicon} and Karl-Olof Lindahl  \href{https://orcid.org/0000-0002-7825-4428}{\orcidicon}} 

\AuthorNames{Björn Lindenberg, Jonas Nordqvist and Karl-Olof Lindahl}

\address[1]{%
Department of Mathematics, Linn\ae us University, 351 95 Växjö, Sweden; jonas.nordqvist@lnu.se~(J.N.); karl-olof.lindahl@lnu.se~(K.-O.L.)
}

\corres{\hangafter=1 \hangindent=1.05em \hspace{-0.82em} Correspondence: bjorn.lindenberg@lnu.se}

\abstract{\textls[-15]{It is well known that ensemble methods often provide enhanced performance in reinforcement learning. In this paper, we explore this concept further by using group-aided training within the distributional reinforcement learning paradigm. Specifically, we propose an extension to categorical reinforcement learning, where distributional learning targets are implicitly based on the total information gathered by an ensemble. We empirically show that this may lead to much more robust initial learning, a stronger individual performance level, and good efficiency on a per-sample~basis.}}


\begin{document}
\section{Introduction}
\label{s:introduction}
The fact that ensemble methods may outperform single agent algorithms in reinforcement learning has been demonstrated numerous times \citep{singh1992efficient,sun1999multi,wiering2008ensemble,fausser2015selective}. These methods can involve combining several algorithms into one agent and then taking actions by a {weighted aggregation scheme} or {rank voting}. However, most conventional ensemble methods in reinforcement learning are often based on expected returns. Perhaps the simplest example is the {average joint policy} derived from an ensemble of independently trained agents, where the action of the ensemble is dictated by the average of the estimated Q-values of each agent.


An alternate view to that of Q-values, the distributional perspective on state-action returns, was discussed in \citep{bellemare2017distributional}. This paradigm represents a shift of focus towards estimating or using underlying distributions of random return variables instead of learning expectations. This in turn paints a complex and more informationally dense picture, and there exists overwhelming empirical evidence that the distributional perspective is helpful in deep reinforcement learning. That is, apart from the possibility of overall stronger performance, algorithmic benefits may also involve the reduction of prediction variance, more robust learning with additional regularization effects, and a larger set of auxiliary goals such as learning risk-sensitive policies \citep{morimura2010parametric, bellemare2017distributional, dabney2018distributional, dabney2018implicit, lyle2019comparative}. Moreover, there have recently been important theoretical works done on understanding the observed improvements and providing theoretical results on convergence~\citep{bellemare2017distributional, rowland2018analysis,lyle2019comparative, bellemare2019distributional}.

In this paper, we propose a group-aided training scheme for distributional
reinforcement learning,
where we merge the distributional perspective with an ensemble method involving agents learning in separate environments. Our main contribution in this regard is the proposed {Ensemble Categorical Control procedure (\algnamn procedure)}. As an initial study, we also provide empirical results where an
\algnamn algorithm
is tested on a subset of Atari 2600 games \citep{bellemare13arcade}, which are standard environments for testing these types of algorithms.

Specifically, \algnamn{} is an extension of {Categorical Distributional Reinforcement Learning} (CDRL), which was introduced in
\citep{bellemare2017distributional} and made explicit in \citep{rowland2018analysis}. Similar to CDRL, we consider distributions 
defined on a fixed discrete support, with projections onto the support for all possible categorical distributions arising internally in the algorithm. For each agent in \algnamn{}, we replace the target generation of CDRL by targets generated by the ensemble mean mixture distribution of the individual target~distributions.

We argue that \algnamn{} implies an implicit sharing of information between agents during learning, where the distributional paradigm gives us more robust targets and an arguably more nuanced aggregated picture, which preserves multimodality. The experiments confirm the validity of the approach, where in all cases, the extension generates strong individual agents and good efficiency when regarded as an ensemble.

The paper is organized in the following way. In Section \ref{s:background}, we give a background to distributional reinforcement learning. In Section \ref{s:learning with ensembles}, we introduce
the proposed \algnamn procedure.
At the end of Section~\ref{s:learning with ensembles}, we give a reference to another contribution of the present work: the pseudocode and source code for an implementation of the \algnamn algorithm. In Section~\ref{s:results}, we present and evaluate the results of our implementation of the \algnamn algorithm on five specific Atari 2600 environments. Finally, in Section \ref{s:discussion}, we~zoom out and discuss the results in a broader context, as well as suggest future work.

\section{Background}
\label{s:background}
We considered agent-environment interactions. For each observed state, the agent selects an action, whereby the environment generates a reward and a next state. Following the framework of~\citep{rowland2018analysis}, we let $\cX$ and $\cA$ denote the sets of states and actions, respectively, and let $p \colon \cX \times \cA \to \sP(\mathbb{R} \times \cX)$ be a {transition kernel} that maps state-action pairs to joint distributions of immediate rewards and next states. Then, we can model this interaction by a {Markov Decision Process} (MDP) $(\cX, \cA, p, \gamma)$, where~$\gamma \in [0,1)$ is a {discount factor} of future rewards. Moreover, an agent can sample its actions through a {stationary policy} $\pi \colon \cX \to \sP(\cA)$, which maps a current state to a distribution over available~actions.

Throughout the rest of this paper, we consider MDPs where $\cX \times \cA$ is a countable state-action space. We denote by $\cD = \sP(\mathbb{R})^{\cX \times \cA}$ the set of functions where $\eta \in \cD$ maps each state-action pair $(x,a)$ to a distribution $\eta^{(x,a)} \in \sP(\mathbb{R})$. Similarly, we put $\cD_n = \sP_n(\mathbb{R})^{\cX \times \cA}$, where $\sP_n(\mathbb{R})$ is the set of probability distributions with finite $n^{\text{th}}$-moments. For a given $\eta \in \cD$, we let $Q_{\eta} \colon \cX \times \cA \to \R$ denote the function that maps state-action pairs $\{(x,a)\}$ to the corresponding first moments of $\{\eta^{(x,a)}\}$, {i.e.},
\begin{equation*}
Q_{\eta}(x,a) \coloneqq \int_{\R} z \ \eta^{(x,a)}(d z).
\end{equation*}

To appreciate a subsequent summary of distributional reinforcement theory fully, we may also need to make the following definition explicit.
\begin{Definition}
For a Borel measurable function $g \colon \R \to \R$ and $\nu \in \sP(\mathbb{R})$, we let $g_\# \nu$ denote the {push-forward measure} defined by:
\begin{equation*}
g_\# \nu(A) \= \nu\left( g^{-1} (A) \right)
\end{equation*}
on all Borel sets $A \subseteq \R$. In particular, given $r, \gamma \in \R$, we let $(f_{r, \gamma})_\# \nu$ be the push-forward measure where $f_{r, \gamma} (x) \= r + \gamma x$.
\end{Definition}

Suppose further that we have a set $\cP$ of categorical distributions supported on a fixed set $\mathbf{z} = \set{z_1, z_2, \dots, z_K}$ of equally-spaced numbers. Then, the following projection operator minimizes the distance between any categorical distribution $\nu = \sum_{i = 1}^n p_i \delta_{y_i}$ and elements in $\cP$ with respect to the {Cramér metric} \citep{rizzo2016energy,lyle2019comparative}.
\begin{Definition}
\label{def:cramer projection}
The {Cramér projection} $\Pi_{\mathbf{z}}$ maps any Dirac measure $\delta_y$ to a distribution in $\cP$ by:
\begin{equation*}
\Pi_{\mathbf{z}}(\delta_y) = \left \{ \begin{array}{ll}
\delta_{z_1} & y \leq z_1, \\
\frac{z_{i+1} - y}{\Delta z} \delta_{z_i} + \frac{y - z_i}{\Delta z} \delta_{z_{i+1}} & z_i < y \leq z_{i+1}, \\
\delta_{z_K} & y > z_K.
\end{array} \right.
\end{equation*}

Moreover, the projection is defined to be linear over mixture distributions such that:
\begin{equation*}
\Pi_{\mathbf{z}}\left(\sum_i p_i \delta_{y_i} \right) = \sum p_i \Pi_{\mathbf{z}}\left(\delta_{y_i}\right).
\end{equation*}
\end{Definition}

\subsection{Expected Reinforcement Learning} Before we go into the distributional perspective, let us first give a quick reminder about some value function fundamentals, here stated in operator form.

Let $(\cX, \cA, p, \gamma)$ be an MDP. Given $(x,a) \in \cX \times \cA$, we define the {return} of a policy $\pi$ as the random variable:
\begin{equation}
\label{eq:return}
Z_\pi(x,a) \= \left. \sum_{t = 0}^\infty \gamma^t R_t \; \middle\vert \; X_0 = x, A_0 = a\right. ,
\end{equation}
where $(R_t)_{t = 0}^\infty$ is a random sequence of immediate rewards, indexed by time step $t$ and dependent on random state-action pairs $(X_t, A_t)_{t = 0}^\infty$ under $p$ and $\pi$.

In an evaluation setting of some fixed policy $\pi$, let $Q_\pi \colon \cX \times \cA \to \mathbb{R}$ be the {expected return function}, which by definition has values:
\begin{equation*}
Q_\pi(x,a) = \E [Z_\pi(x,a)].
\end{equation*}

If we consider distributions dictated by $p$ and $\pi$ and let $R(x,a)$ and $(X',A')$ denote the random reward and subsequent random state-action pair given $(x,a) \in \cX \times \cA$, then we recall the {Bellman operator} $\cT^\pi$ defined by:
\begin{equation}
\label{eq:bellman}
\forall (x,a) \ \left(\cT^\pi g\right)(x,a) = \E_{p} [ R(x,a)] + \gamma \E_{p, \pi}[g(X',A')]
\end{equation}
on bounded real functions $g \in \cB(\cX \times \cA, \R)$. Moreover, in the search for values attained by optimal policies, we also recall the {optimality operator} $\cT^*$ where:
\begin{equation}
\label{eq:optimal}
\forall (x,a) \ \left(\cT^* g\right)(x,a) = \E_{p} [ R(x,a)] + \gamma \E_{p}[\max_{a'} g(X',a')].
\end{equation}

It is readily verified that both operators are contraction maps on the complete metric space $\left(\cB(\cX \times \cA, \R), d_\infty\right)$. In addition, their unique fixed points are given by $Q_\pi$ and $Q^*$, respectively, where~$Q^*$ is the optimal function defined by:
\begin{equation*}
Q^*(x,a) = \max_{\pi} Q_\pi(x,a)
\end{equation*}
for all $(x,a)$ \citep{bertsekas1996neuro}.

\subsection{Distributional Reinforcement Learning}
We now proceed by presenting some of the main ideas of distributional reinforcement learning in a tabular setting. We will first look at the {evaluation problem}, where we are trying to find the state-action value of a fixed policy $\pi$. Second, we consider the {control problem}, where we try to find the optimal state-action value. Third, we consider the {distributional approximation procedure} CDRL used by agents in this paper.

\subsubsection{Evaluation} 

We consider a distributional variant of \eqref{eq:bellman}, the {distributional Bellman operator} given by \mbox{$T^\pi \colon \cD \to \cD$, }
\begin{equation}
\label{eq:dbellman}
\forall (x,a) \ \left(T^\pi \eta\right)^{(x, a)} \= \int_\mathbb{R} \sum_{(x',a') \in \cX \times \cA} (f_{r, \gamma})_\# \eta^{(x',a')} \ \pi(a' \mid x') p(d r,x'\mid x,a).
\end{equation}

Here, $T^\pi$ is, for all $n \geq 1$, a $\gamma$-contraction in $\cD_n$ with a unique fixed point when $\cD_n$ is endowed with the supremum $n^{\text{th}}$-Wasserstein metric (\citep{bellemare2017distributional}, Lemma~3) (see \citep{villani2008optimal} for more details on Wasserstein distances). Moreover by Proposition 2 of \citep{lyle2019comparative}, $T^\pi$ is expectation preserving when we have an initial coupling with the $\cT^\pi$-iteration given in \eqref{eq:bellman}; that is, given an initial $\eta_0 \in \cD$ and a function $g$, such that $g = Q_{\eta_0}$. Then, $\left(\cT^\pi\right)^n g = Q_{\left(T^\pi\right)^n \eta_0}$ holds for all $n\geq 0$.

Thus, if we let $\eta_\pi \in \cD$ be the function of distributions of $Z_\pi$ in \eqref{eq:return}, then $\eta_\pi$ is the unique fixed point satisfying the {distributional Bellman equation:}
\begin{equation*}
\eta_\pi = T^\pi \eta_\pi.
\end{equation*}

It follows that iterating $T^\pi$ on any starting collection $\eta_0$ with bounded moments eventually solves the evaluation task of $\pi$ to an arbitrary degree.

\subsubsection{Control} Recall the Bellman optimality operator $\cT^*$ of \eqref{eq:optimal}. If we define a corresponding {distributional optimality operator} $T^* \colon \cD \to \cD$,
\begin{equation}
\label{eq:doptimal}
\forall (x,a) \ (T^* \eta)^{(x,a)} \= \int_\mathbb{R} \sum_{(x',a') \in \cX \times \cA} (f_{r, \gamma})_\# \eta^{(x',a^*(x'))} \ p(d r,x'\mid x,a),
\end{equation}
where $a^*(x') = \argmax_{a' \in \cA} Q_{\eta}(x',a')$, then expectation values generated by iterates under $T^*$ will behave as expected. That is, if we put $Q_n \coloneqq Q_{(T^*)^n \eta_0}$, then we have an exponentially fast uniform convergence $Q_n \to Q^*$ as $n \to \infty$. However, $T^*$ is not a contraction in any metric over distributions and may lack fixed points altogether in $\cD$ \citep{bellemare2017distributional}.

\subsubsection{Categorical Evaluation and Control}
In most real applications, the updates of \eqref{eq:dbellman} and \eqref{eq:doptimal} are either computationally infeasible or impossible to fully compute due to $p$ being unknown. It follows that approximations are key to defining practical distributional algorithms. This could involve {parametrization} over some selected set of distributions along with {projections} onto these distributional subspaces. It could also involve {stochastic approximations} with sampled transitions and gradient updates with {function approximation}.

A structure for algorithms making use of such approximations is {Categorical Distributional Reinforcement Learning} (CDRL). In what follows is a short summary of the CDRL procedure fundamental to single agent implementations in this paper.

Let $\mathbf{z} = \set{z_1, z_2, \dots, z_K}$ be an ordered fixed set of equally-spaced real numbers such that $z_1 < z_2 < \cdots < z_K$ with $\Delta z \= z_{i+1} - z_i$. Let:
\begin{equation*}
\cP = \left \{ \sum_{i = 1}^K p_i \delta_{z_i} : p_1, \dots, p_K \geq 0, \ \sum_{i = 1}^K p_i = 1 \right \} \subset \sP(\R)
\end{equation*}
be the subset of categorical distributions in $\sP(\R)$ supported on $\mathbf{z}$. We consider parameterized distributions by using $\whcD = \cP^{\cA \times \cX}$ as the collection of possible inputs and outputs of an algorithm. Moreover, for each $\eta \in \whcD$, we have:
\begin{equation*}
Q_{\eta}(x, a) = \sum_{i = 1}^K p_i(x, a) z_i.
\end{equation*}
as its Q-value function.

Given a subsequent treatment of our extension of CDRL, we first reproduce the steps of the general procedure in Algorithm~\ref{alg:CDRL} (see \citep{rowland2018analysis}, Algorithm~1).
\begin{algorithm}
\caption{Categorical Distributional Reinforcement Learning (CDRL) \label{alg:CDRL}}
\begin{enumerate}
\item \label{alg:sample} At each iteration step $t$ and input $\eta_t \in \whcD$, sample a transition $(x_t, a_t, r_t, x'_t)$.
\item \label{alg:action} Select $a^*$ to be either sampled from $\pi(x_t)$ in the evaluation setting or taken as $a^* = \argmax_a Q_{\eta_t}(x'_t, a)$ in the control setting.
\item \label{alg:target} Recall the Cramér projection $\Pi_{\mathbf{z}}$ given in Definition~\ref{def:cramer projection}, and put:
\begin{equation*}
\wheta_t^{(x_t,a_t)} \coloneqq \Pi_{\mathbf{z}} \left(f_{r_t}\right)_\# \eta_t^{(x'_t,a^*)}.
\end{equation*}
\item \label{alg:update} Take the next iterated function as some update $\eta_{t+1}$ such that:
\begin{equation*}
\KL\left(\wheta_t^{(x_t,a_t)} \parallel \eta_{t+1}^{\left(x_t,a_t\right)} \right) < \KL\left(\wheta_t^{(x_t,a_t)} \parallel \eta_{t}^{\left(x_t,a_t\right)} \right),
\end{equation*}
where:
\begin{equation*}
\KL(\mathbf{p} \parallel \mathbf{q}) \coloneqq \sum_{i = 1}^K p_i \log \left(\frac{p_i}{q_i}\right)
\end{equation*}
denotes the Kullback--Leibler divergence.
\end{enumerate}
\end{algorithm}


Consider first a finite MDP and a tabular setting. Define $\wheta_t^{(x,a)} \= \eta_t^{(x,a)}$ whenever $(x,a) \neq (x_t, a_t)$. Then, by the convexity of $-\log(z)$, it is readily verified that updates of the form:
\begin{equation*}
\eta_{t+1} = (1-\alpha_t)\eta_{t} + \alpha_t \wheta_t \qquad (\alpha_t \in (0,1))
\end{equation*}
satisfy Step~\ref{alg:update}. In fact, if there exists a unique policy $\pi^*$ associated with the convergence of \eqref{eq:optimal}, then~this update yields an almost sure convergence, with respect to the supremum-Cramér metric, to~a~distribution in $\whcD$ with $\pi^*$ as the greedy policy (with some additional assumptions on the stepsizes $\alpha_t$ and sufficient support (see \citep{rowland2018analysis}, Theorem~2, for details).

In practice, we are often forced to use function approximation of the form:
\begin{equation*}
\eta^{(x,a)} = \phi(x,a;\btheta),
\end{equation*}
where $\phi$ is parameterized by some set of weights $\btheta$. Gradient updates with respect to $\btheta$ can then be made to minimize the loss:
\begin{equation}
\label{eq:kl function approximation}
\KL\left(\wheta_t^{(x_t, a_t)} \parallel \phi\left(x_t,a_t;\btheta\right) \right),
\end{equation}
where $\wheta_t^{(x_t, a_t)} = \Pi_{\mathbf{z}} \left(f_{r_t}\right)_\# \phi(x'_t,a^*;\btheta_{\text{fixed}})$ is the computed learning target of the transition $(x_t, a_t, r_t, x'_t)$.
However convergence with the Kullback--Leibler loss and function approximation is still an open question. Theoretical progress has been made when considering other losses, although we may lose the stability benefits coming from the relative ease of minimizing \eqref{eq:kl function approximation} \citep{bellemare2017cramer, bellemare2019distributional, lyle2019comparative}.

An algorithm implementing CDRL with function approximation is \texttt{C51}
\citep{bellemare2017distributional}. It essentially uses the same neural network architecture and training procedure as DQN
\citep{mnih2015human}. To increase stability during training, this also involves sampling transitions from an {experience buffer} and maintaining an older, periodically updated, copy of the weights for target computation. However, instead of estimating Q-values, \texttt{C51} uses a finite support $\mathbf{z}$ of $51$ points and learns discrete probability distributions $\phi(x,a;\btheta)$ over $\mathbf{z}$ via soft-max transfer. Training is done by using the KL-divergence as the loss function over batches with computed targets $\wheta^{(x,a)}$ of CDRL.

\section{Learning with Ensembles}
\label{s:learning with ensembles}
\vspace{-9pt}
\subsection{Ensembles}
Ensemble methods have been widely used in both {supervised learning} and reinforcement learning. In supervised learning, this can involve {bootstrap aggregating} predictors for better accuracy when given unstable processes such as neural networks or using ``expert'' opinion mixtures for better estimators \citep{breiman1996bagging, goodfellow2016deep}. A simple example that demonstrates the possible benefits of aggregation is the following {average pool} of $k$ regression models: Given a sample to predict, assume that the models draw prediction errors $\e_i$, $i = 1, \dots, k$ from a zero-mean multivariate normal distribution with $\E[\e_i^2] = \sigma^2$ and correlations $\rho_{i j} = \rho$. Then, the error made by averaging their predictions is $\e \= (1/k) \sum_{i = 1}^k \e_i$ with:
\begin{equation*}
\E[\e^2] = \left(1 + \rho(k-1)\right)\frac{\sigma^2}{k}.
\end{equation*}

It follows that the mean squared error goes to $\sigma^2/k$ as $\rho \to 0$, whereas we get $\sigma^2$ and no benefit when the errors are perfectly correlated.

Under the assumption of independently trained agents, we have a reinforcement learning variant of the average pool in the following definition.
\begin{Definition}
\label{def:avg}
Given an ensemble of $k$ agents, let $\whQ^{(i)}$ denote the Q-value function estimate of agent $i$, and let $\whQ \coloneqq (1/k) \sum_{i=1}^k \whQ^{(i)}$ denote the mean function. Then, the {average joint policy} $\overline{\pi}$ selects actions according to:
\begin{equation*}
a^* = \argmax_a \whQ(x,a) = \argmax_a \frac{1}{k} \sum_{i = 1}^k \whQ^{(i)}(x,a).
\end{equation*}
at every $x \in \cX$.
\end{Definition}

Thus, $\overline{\pi}$ represents an aggregation strategy where we consider the information provided by each agent as equally important. Moreover, by the linearity of expectations and in view of \eqref{eq:optimal}, if~we have initial functions $Q_0^{(i)}$ with $n$-step ensemble values $Q_n \=(1/k) \sum_{i = 1}^k Q_n^{(i)}$, then full updates $Q_{n}^{(i)} \= \cT^* Q_{n-1}^{(i)}$ of each agent will yield $Q_{n} = \cT^* Q_{n-1}$ for the ensemble. Assume further that learning is done with a single algorithm in separate environments. If we take $\whQ^{(i)}(x,a)$ as estimates of $Q_n^{(i)}(x,a)$ for some step $n$, with errors $\e_i$ distributed as multivariate Gaussian noise, then we should expect $\whQ(x,a)$ to have a smaller expected error variance in its estimation of $Q_n(x,a)$ similar to regression models. This implies more robust performance when given an unstable training process far from convergence, but it also implies diminishing improvements when the algorithm is close to converging to a unique policy.

\begin{figure}[H]
\centering
\includegraphics[width=0.95\textwidth]{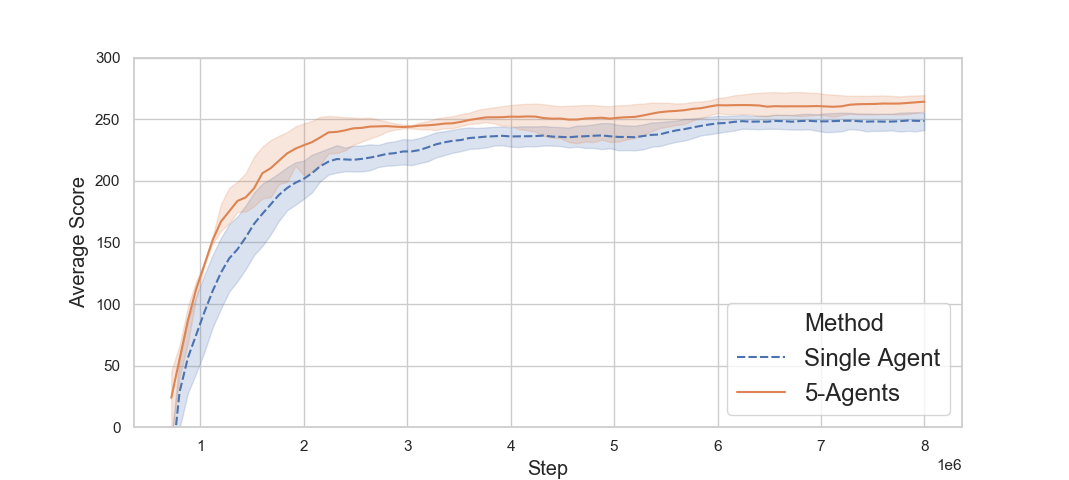}
\caption{Low capacity CDRL implementations in the LunarLander-v2 environment. We can see that the enhanced performance of an average joint policy of five agents may not vanish due to agents settling on distinct sub-optimal policies.}
\label{fig:lunar} 
\end{figure}

However, in real applications, and in particular with function approximation, there may be instances where the improved performance by $\overline{\pi}$ does not vanish due to agents converging to distinct sub-optimal policies. An illustration of this phenomenon can be seen in Figure~\ref{fig:lunar}. It shows evaluations during learning in the \texttt{LunarLander-v2} environment \citep{openai}. The single agents used CDRL on a 29 atom support. To approximate distributions, the agents used small neural networks with three encoding layers consisting of 16 units each. The architecture was purposely chosen to make it harder for the optimizer to converge to an optimal policy, possibly due to lack of capacity. At each evaluation point, the models were tested with $\e = 0.001$. The figure also includes evaluations of average joint policies of five agents having the same evaluation $\e$. However, we can see that the joint information provided by an ensemble of five agents transcends individual capacity, indicating that some agents settle on distinct sub-optimal solutions.

\subsection{Ensemble Categorical Control}
\label{sec:catens}
We consider an ensemble of $k$ agents, each independently trained with the same distributional algorithm, where $\eta_{i}$, $i = 1, \dots, k$ are their respective distributional collections. There are several ways to aggregate distributional information provided by the ensemble with respect to forecasts and risk-sensitivity \citep{clemen1999combining,casarin2016bayesian}. Perhaps the simplest is a distributional variant of the average joint policy, where we consider the mean function $\overline{\eta}$ of mixture distributions:
\begin{equation}
\label{eq:meanmix}
\forall (x,a) \ \overline{\eta}^{(x,a)} \= \frac{1}{k} \sum_{i = 1}^k \eta_{i}^{(x,a)}.
\end{equation}

Since $\overline{\eta}^{(x,a)}$ is a linear pool, it preserves multimodality during aggregation. Hence, it maintains an arguably more nuanced picture of estimated future rewards compared to methods that generate unimodal aggregations around unrealizable expected values. In addition, expectations under $\overline{\eta}$ yield the Q-function used by the average joint policy in Definition~\ref{def:avg} with all the performance benefits that this entails during learning.

The finite support of the CDRL procedure may provide another reason to aggregate by $\overline{\eta}$: Under the assumption that $\eta_{i}^{(x,a)}$, $i = 1, \dots, k$ are drawn as random vectors from some multivariate normal population with mean $\bm{\mu}(x,a)$ and covariance $\bm{\Sigma}(x,a)$, then $\overline{\eta}$ is a maximum likelihood estimate of the mean categorical distribution $\bm{\mu}(x,a)$ induced by the algorithm over all possible training runs~\citep{johnson2014applied}. It~follows that $\overline{\eta}$ may provide more robust estimates in reflecting mean $t$-step capabilities of the procedure in terms of distributions found by sending $k \to \infty$.

It then stands to reason that \eqref{eq:meanmix} should help accelerate learning by providing better and more robust targets in the control setting of CDRL. This implies implicitly sharing information gained between agents and following accelerated learning trajectories closer to the true expected capability of an algorithm. We can summarize this as an extension of the CDRL control procedure.

For a fixed support $\mathbf{z}$, we parameterize individual distribution functions $\eta_{i,t}$, $i=1,\dots,k$, at time step $t$ by using $\whcD = \cP^{\cA \times \cX}$ as possible inputs and outputs of the algorithm. Let $\overline{\eta}_t$ be the mean function of $\{\eta_{i,t}\}_{i = 1}^k$ according to \eqref{eq:meanmix}. The extension is then given by Algorithm~\ref{alg:ecc procedure}.
\begin{algorithm}
\caption{Ensemble Categorical Control (\algnamn) \label{alg:ecc procedure}}
\begin{enumerate}
\item\label{ens:sample} At each iteration step $t$ and for each agent input $\eta_{i,t}$, sample a transition $(x,a,r,x')$.
\item \label{ens:action} Let $a^* = \argmax_{a'} Q_{\overline{\eta}_t}(x',a')$.
\item \label{ens:target} Recall the Cramér projection $\Pi_{\mathbf{z}}$ given in Definition~\ref{def:cramer projection}, and put:
\begin{equation*}
\wheta_{i,t}^{(x,a)} \coloneqq \Pi_{\mathbf{z}} \left(f_{r}\right)_\# \overline{\eta}_t^{(x', a^*)}.
\end{equation*}
\item\label{ens:update} For each agent, follow Step~\ref{alg:update} of CDRL with target $\wheta_{i,t}^{(x,a)}$.
\end{enumerate}
\end{algorithm}


We note that if updates are done in full or on the same transitions, then the algorithm trivially reduces to CDRL by the linearity of $(f_{r})_\#$; hence, we lose the benefits of the ensemble.

To avoid premature convergence to correlated errors, we would ideally want the agents to have the freedom to explore different trajectories during learning. In the case of function approximation, this can involve maintaining a separate experience buffer for each agent. It can also involve periodical updates of {ensemble target networks} in the hope of generating sufficiently diverse policies until convergence. The latter is in practical terms the only way to minimize overhead costs induced by inter-thread ensemble queries in simulations. Too short periods here imply fast initial learning; but with correlated errors, high overhead costs, and instability \citep{mnih2015human}. Long periods would imply the possibility of more diverse policies, but with slower learning. The pseudocode for an algorithm using function approximation with \algnamn{} can be found in Algorithm \ref{alg:ecc}. The source code for an implementation of \algnamn{} can be found at \cite{bjorn_lindenberg_2020_3742688}.

\section{Empirical Results on a Subset of Atari 2600 Games}
\label{s:results}
As a first step in understanding the properties of the extension \algnamn{} discussed in Section~\ref{sec:catens}, we now evaluate an implementation of the procedure on five Atari 2600 environments found in the {Arcade Learning Environment} \citep{bellemare13arcade, openai, stable-baselines}.

Specifically, we looked at ensembles of $k = 5$ agents. To get a proper comparison of the algorithms, we employed for all agents the well-tested architecture, hyperparameters, and training procedure as \texttt{C51} in \citep{bellemare2017distributional}; except for a slightly smaller individual replay buffer size at 900 K. This yielded an implicit buffer size of 4.5 M for the entire \algnamn ensemble. In addition, we employed for each \algnamn{} agent a larger ensemble target network. The network consisted of copied weights from all \algnamn networks and was updated periodically at every 10K steps with negligible overhead.

We trained $k$ agents on the first 40 M frames (roughly 185 h of Atari-time at 60 Hz). Agent models were saved every 400 K frames. For each save, we evaluated the performance of the individual agents (\algnamn agent) and the ensemble with an average joint policy (\algnamn ensemble). Moreover, we took an ensemble of $k = 5$ independently trained agents using $\overline{\pi}$ as our baseline (CDRL joint). For comparison, we also evaluated each such single agent (CDRL agent). In all performance protocols, we started an episode under the 30 no-op regime \citep{mnih2015human} with an exploration epsilon set to $\e = 0.001$. The evaluation period was 500 K frames with episodes truncated at 108 K frames (30 min).

In our particular implementation in \cite{bjorn_lindenberg_2020_3742688}, each algorithm required roughly two days of compute time per environment for training and evaluation combined. Single replay buffers used \textasciitilde 35 GB of optimized RAM (\textasciitilde 47 GB raw); hence, we used \textasciitilde 175 GB of RAM for concurrently training the \algnamn ensemble.
\subsection{Online Performance}

\textls[-5]{To get a sense of the algorithmic robustness and speed of learning, we report the {online performance} of agents and ensembles \citep{dabney2018distributional}. Under this protocol, we recorded the average return for each evaluation point during learning. We also stored the best average return score for all points of each seed.}

\begin{figure}[H]
\centering
\includegraphics[width=0.95\textwidth]{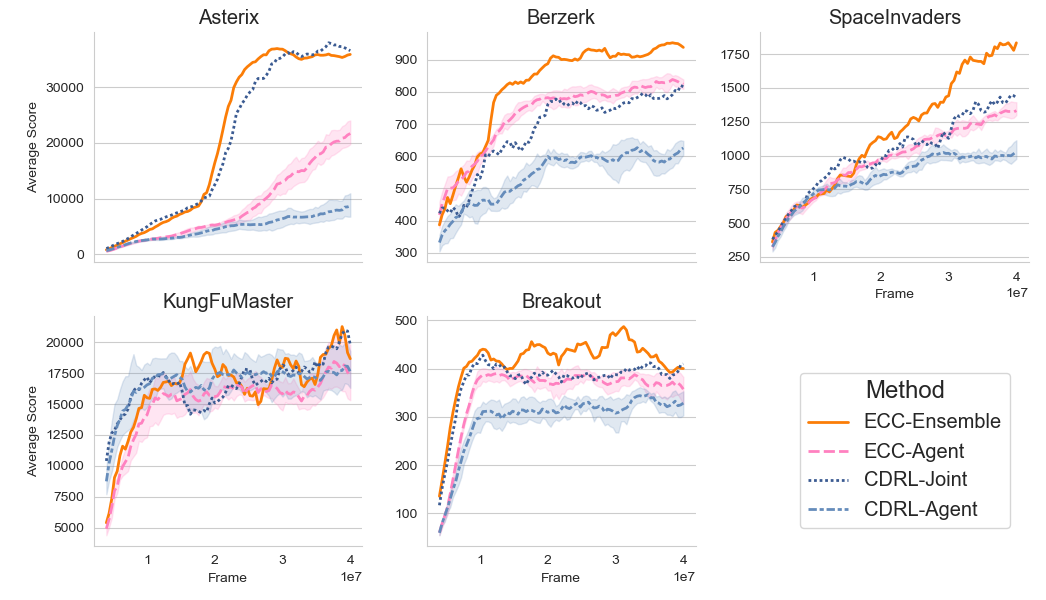}
\caption{Online performance over the first 40 M frames. The evaluation scores shown are moving averages over 4 M frames. The data are available at \cite{bjorn_lindenberg_2020_3742688}. }
\label{fig:online}
\end{figure}

We can see in Table~\ref{tab:best} and Figure~\ref{fig:online} that the extension ensemble was on par or outperformed the baseline in online performance over all five environments. Moreover, in four out of five games, single~\algnamn{} agents had similar performance to the joint policy of $k$ independently trained agents, which~was the main training objective of the extension algorithm. We also note that in all environments, except possibly \texttt{Breakout} and \texttt{KungFuMaster}, ensemble agents seemed to be uncorrelated enough to generate a boost in performance by their joint information, while \algnamn{} agents had a better individual performance than single CDRL agents in four out of five games.
\unskip

\begin{table}[H]
\centering
\caption{Best achieved evaluation scores in online performance over the first 40 M samples, here with 95\% confidence when there is more than one seed. The data are available at \cite{bjorn_lindenberg_2020_3742688}.}
\label{tab:best}
\begin{tabular}{ccccc}
\toprule
\textbf{Game} & \textbf{CDRL Agent} & \textbf{\algnamn Agent} & \textbf{CDRL Joint} & \textbf{\algnamn Ensemble} \\
\midrule
Asterix & \numprint{12998} $\pm$ {3042} & \numprint{28196} $\pm$ \numprint{903}& \numprint{39413} & \numprint{38938} \\
Berzerk & \npr{795}{47} & \npr{958}{12} & \numprint{890}& {1034} \\
SpaceInvaders & {1429} $\pm$ {91} & {1812} $\pm$ {87} & {1850} & {2395}\\
Breakout & \npr{444}{44} & \npr{546}{27} & \numprint{515} & \numprint{665} \\
KungFuMaster & {27,984} $\pm$ {1767}& {27,302} $\pm$ {2213}& \numprint{25826} & \numprint{29629} \\
\bottomrule
\end{tabular} 

\end{table}

\subsection{Relative Ensemble Sample Performance}
Although ensembles will digest frames at nearly $k$ times the rate of a single CDRL algorithm, we considered here the {relative sample performance}, where we looked at performance versus the total information accumulated by an algorithm. Under this protocol, we measured the relative ratio of mean evaluation scores as a function of the {total amount of frames} seen by each learning system. This would give us an idea of how efficiently an ensemble algorithm could translate experience into performance on a per-sample basis compared to single CDRL. Note that if single CDRL agents all converged to correlated errors, then the joint policy should eventually converge to $1/k$-efficiency in relative sample performance. Thus, in general, we should expect the relative performance to degrade as training progresses with diminishing ensemble benefits.

Table~\ref{tab:rel} shows the measured relative performance of the two ensemble methods, averaged over the first 40 M samples. We note that initial learning with ensembles may generate performance much higher than $1/k$-efficiency. We also note that the extension ensemble came close to full efficiency in \texttt{Berzerk} and \texttt{Breakout}, {i.e.}, it displayed a near $k$-factor increase in learning rate. However, depending on the environment, the actual speed-up may vary wildly during learning, as shown in Figure~\ref{fig:online}.
\begin{table}[H]
\centering
\caption{Rough estimates of relative sample performance, here expressed as percentages of CDRL agent performance and averaged over the first 40 M samples. The data are available at \cite{bjorn_lindenberg_2020_3742688}.}
\label{tab:rel}
\begin{tabular}{cccccc}
\toprule
\textbf{Method} & \textbf{Asterix} & \textbf{Berzerk }& \textbf{Breakout} & \textbf{SpaceInvaders} & \textbf{KungFuMaster}\\
\midrule
\algnamn Ensemble & 47.7 \% & 93.7 \% & 93.7 \% & 63.4 \% & 66.9 \% \\
CDRL Joint & 56.3 \% & 86.7 \% & 86.1 \% & 67.2 \% & 87.0 \% \\
\bottomrule
\end{tabular}

\end{table}

\section{Discussion}
\label{s:discussion}
In this paper, we proposed and studied an extension of categorical distributional reinforcement learning, where we employed averaged learning targets over an ensemble. This extension implied an implicit sharing of information between agents during learning, where under the distributional paradigm, we should expect a richer and more robust set of predictions while preserving multimodality during aggregation. To test these assumptions, we did an initial empirical study on a subset of Atari 2600 games, where we employed essentially the same architecture and hyperparameter set as the \texttt{C51} algorithm in \citep{bellemare2017distributional}. In all cases, we saw that the single agent performance objective of the extension was accomplished. We also studied the effects of keeping extension amplified agents in an ensemble, where in some cases, the performance benefits were present and stronger than an averaged ensemble of independent agents.

\textls[-15]{We note that unlike massively distributed approaches such as Ape-X \citep{horgan2018distributed}, the extension represents a decentralized distributed learning system with minimal overhead. As such, it naturally comes with poor scalability, but with greater efficiency on a per-sample basis. An interesting idea here would be to somewhat counteract the poor scalability by choosing agents with successively lower capacity as the ensemble size increases. We should then expect to see better performance with increasing size until a cutoff point is reached, hinting at the minimum capacity needed to find and represent strong solutions~effectively.}

We leave as future work the matter of convergence analysis and hyperparameter tuning, in~particular the update period for a target ensemble network. It is quite possible that the update frequency of \texttt{C51} was too aggressive when using ensemble targets. This may lead to premature convergence to correlated agents upon reaching difficult environmental plateaus with rarely seen transitions to more abundant rewards. Some interesting ideas here would be scheduled update periods or eventually switching to CDRL from a much stronger and robust level of individual performance. However, to gauge these matters fully, we would need a more comprehensive empirical study.
\vspace{6pt}



\authorcontributions{Conceptualization, B.L., J.N., and K.-O.L.; methodology, B.L. and J.N.; software, B.L.; validation, B.L.; formal analysis, B.L., J.N., and K.-O.L.; investigation, B.L.; data curation, B.L.; writing, original draft preparation, B.L.; writing, review and editing, B.L., J.N., and K.-O.L.; visualization, B.L.; supervision, K.-O.L.; project administration, K.-O.L. All authors read and agreed to the published version of the manuscript.}

\funding{This research received no external funding.}

\acknowledgments{The authors would like to thank the referees for comments that helped improve the presentation. The authors would also like to thank Morgan Ericsson, {Department of Computer Science and Media Technology, Linn\ae us University}, for productive discussions and technical assistance with the {LNU-DISA
High Performance Computing Platform}. }

\conflictsofinterest{The authors declare no conflict of interest.}


\newpage

\abbreviations{The following abbreviations are used in this manuscript:\\

\noindent
\begin{tabular}{@{}ll}
CDRL & Categorical Distributional Reinforcement Learning\\
MDP & Markov Decision Process\\
ECC & Ensemble Categorical Control
\end{tabular}}
\appendixtitles{no} 
\appendix
\section{} \label{appendixA}

\setcounter{algocf}{0}
\renewcommand\thealgocf{A\arabic{algocf}}
\setlength{\algotitleheightrule}{0.5pt} 
\begin{algorithm}[H]
\caption{Ensemble categorical control. \label{alg:ecc}}
\setstretch{1.5}
\begin{algorithmic}
\REQUIRE Number of iteration steps $N$, ensemble size $k$, support $\mathbf{z}$
\STATE Initialize starting states $x_1, \dots, x_k$ in independent environments
\STATE Initialize agent networks $\eta_{\btheta_1}, \dots, \eta_{\btheta_k}$ with random parameters $\btheta_1, \dots, \btheta_k$
\STATE Initialize target network $\overline{\eta} = \frac{1}{k}\sum_i \eta_{\btheta_i^-}$ with $\btheta_i^- \leftarrow \btheta_i$
\STATE Initialize replay buffers $\cB_1, \dots, \cB_k$ with the same size $S$
\FOR{$t=1$ \TO $N$}
\FORALL{$i \in \set{1, \dots, k}$}
\STATE Set $a_i$ to be a uniform random action with probability $\e_t$
\STATE Otherwise, set $a_i \leftarrow \argmax_{a'} Q_{\eta_{\btheta_i}}(x_i, a')$
\STATE Execute $a_i$, and store the transition $(x_i, a_i, r_i, x_i')$ in $\cB_i$
\STATE Set $x_i \leftarrow x_i'$
\ENDFOR
\IF{$t\equiv 0 \mod P_{\text{update}}$}
\FORALL{$i \in \set{1, \dots, k}$}
\STATE Initialize loss $L \leftarrow 0$
\STATE Sample uniformly a minibatch $B \subset \cB_i$
\FORALL{$(x,a,r,x') \in B$}
\STATE Set $a^* \leftarrow \argmax_{a'} Q_{\overline{\eta}}(x',a')$
\STATE Set $L \leftarrow L +\KL\left(\Pi_{\mathbf{z}} \left(f_{r}\right)_\# \overline{\eta}^{(x', a^*)} \parallel \eta_{\btheta_i}^{(x,a)}\right)$
\ENDFOR
\STATE Update $\btheta_i$ by a gradient descent step on $L$
\ENDFOR
\ENDIF
\IF{$t\equiv 0 \mod P_{\text{clone}}$}
\FORALL{$i \in \set{1, \dots, k}$}
\STATE Update target network with $\btheta_i^- \leftarrow \btheta_i$
\ENDFOR
\ENDIF
\ENDFOR
\end{algorithmic}
\end{algorithm}



\reftitle{References}

\end{document}